\documentclass[letterpaper, 10 pt,
conference]{ieeeconf} %

\IEEEoverridecommandlockouts %

\usepackage{graphics} %
\usepackage{graphicx}
\usepackage{subfig}
\usepackage{amsmath} %
\usepackage{amssymb} %
\usepackage{xcolor} \usepackage{multirow} \usepackage{algorithm}
\usepackage{algpseudocode} \usepackage[font={small}]{caption}

\algrenewcommand\algorithmicindent{0.5em}%

\DeclareMathOperator*{\argmax}{arg\,max}

\title{\LARGE \bf Modeling Cooperative Navigation in Dense Human
  Crowds }

\author{Anirudh Vemula$^{1}$, Katharina Muelling$^{1}$ and Jean
  Oh$^{1}$%
  \thanks{$^{1}$A. Vemula, K. Muelling and J. Oh are with the Robotics
    Institute, Carnegie Mellon University, Pittsburgh, Pennsylvania,
    USA 15213.  Email: \texttt{avemula1@andrew.cmu.edu},
    \texttt{\{kmuelling, jeanoh\}@nrec.ri.cmu.edu}}%
}

\begin{document}

\newcommand{\fbold}{\mathbf{f}}
\newcommand{\supi}{^{(i)}}
\newcommand{\supj}{^{(j)}}
\newcommand{\zbold}{\mathbf{z}}
\newcommand{\subt}{_{1:t}}
\newcommand{\supR}{^{(R)}}
\newcommand{\Gbold}{\mathbf{G}}
\newcommand{\Obold}{\mathbf{O}}
\newcommand{\red}{\textcolor{red}}
\newcommand{\xbold}{\mathbf{x}}
\newcommand{\gbold}{\mathbf{g}}
\newcommand{\Xbold}{\mathbf{X}}
\newcommand{\xdbold}{\mathbf{x'}}
\newcommand{\ybold}{\mathbf{y}}
\newcommand{\xbvel}{\frac{\Delta \xbold}{\Delta t}}
\newcommand{\ybvel}{\frac{\Delta \ybold}{\Delta t}}
\newcommand{\xvel}{\frac{\Delta x}{\Delta t}}
\newcommand{\yvel}{\frac{\Delta y}{\Delta t}}
\newcommand{\subH}{_{t+1:t+H}}
\newcommand{\subti}{_{1:t-1}}
\newcommand{\xbvelinb}{\xbvel^{(i \in B_g)}}
\newcommand{\subtplus}{_{t+1}}
\newcommand{\supP}{^{(P)}}

\maketitle
\thispagestyle{empty}
\pagestyle{empty}

\begin{abstract}
For robots to be a part of our daily life, they need to be
able to navigate among crowds not only safely but also in a
socially compliant fashion. This is a challenging problem
because humans tend to navigate by implicitly cooperating
with one another to avoid collisions, while heading toward
their respective destinations. Previous approaches have
used hand-crafted functions based on proximity to model
human-human and human-robot interactions. However, these
approaches can only model simple interactions and fail to
generalize for complex crowded settings. In this paper, we
develop an approach that models the joint distribution over
future trajectories of all interacting agents in the crowd,
through a local interaction model that we train using real
human trajectory data. The interaction model infers the
velocity of each agent based on the spatial orientation of
other agents in his vicinity. During prediction, our
approach infers the goal of the agent from its past
trajectory and uses the learned model to predict its future
trajectory. We demonstrate the performance of our method
against a state-of-the-art approach on a public dataset and
show that our model outperforms when predicting
future trajectories for longer horizons. 
\end{abstract}

\section{INTRODUCTION}
\label{sec:introduction}

There is an increasing need for robots to operate in the midst of
human crowds. This requires the robot to be able to navigate through a
crowd in a socially compliant way, i.e., the robot needs to
collaboratively avoid collisions with humans and adapt its
trajectories in a human predictable manner.
To date, the majority of existing works in the area of social
navigation has focused on the prediction of individual motion patterns
in the crowd to improve the navigation performance~\cite{thompson09,
  bennewitz05, large04}. However, even in the case of perfect
prediction, these approaches can lead to
severely suboptimal paths \cite{trautman10}.
The primary reason for such underperformance is that these approaches
do not capture the complex and often subtle interactions that take
place among humans in a crowd; that is, these approaches model each
agent independently of the others. This observation leads to the
insight that agents in crowds engage in joint collision avoidance.

\begin{figure}[t!]
  \centering
  \includegraphics[width=\linewidth]{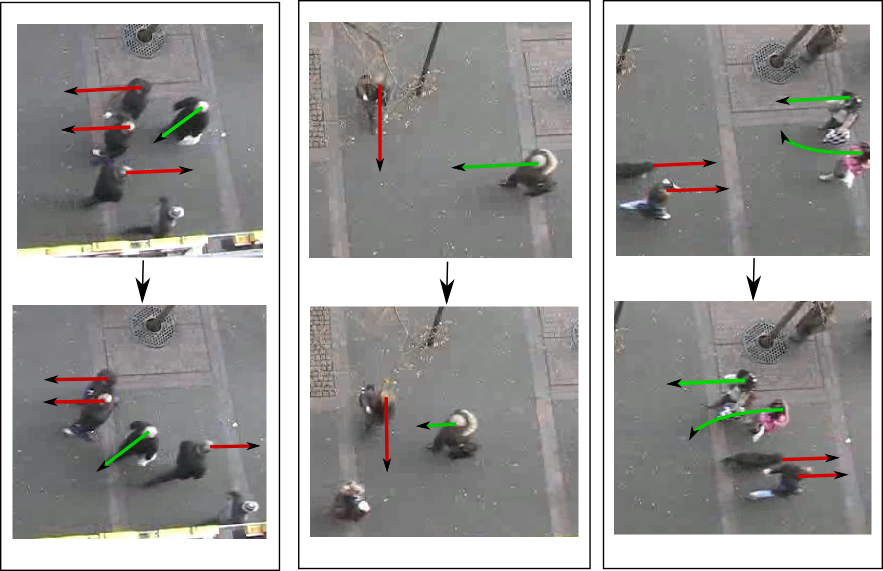}
  \caption{Examples of pedestrians exhibiting cooperative behavior. In
    each image, the velocity of the pedestrian is shown as an arrow
    where the length of each arrow represents the speed. (Left) The
    pedestrian (with green arrow) anticipates the open space between
    the other three (with red arrow) and doesn't slow down. (Middle)
    The pedestrian (with green arrow) slows down to allow the other
    pedestrian (with red arrow) to pass through. (Right) The two
    pedestrians (with green arrows) make way for the oncoming agents
    (with red arrow) by going around them.}
  \label{fig:intro}
  \vspace{-15pt}
\end{figure}

Humans navigate through dense crowds by adapting their trajectories
to those of other people in the vicinity.
Figure~\ref{fig:intro} shows three examples of such behavior where
people pass through, slow down, or go around when they are near
other pedestrians.
In order to learn to navigate in a socially compliant way, it is key
to capture such human-human interactions observed in a crowd.
Pioneering works by~\cite{helbing95,hall63} propose hand-crafted
functions to model such interactions based on proximity. Such
functions are, however, limited in the complexity of interactions that
they can model and fail to generalize for crowded settings. Trautman
et. al. \cite{trautman10} proposed an approach that explicitly models
human-human and human-robot interactions to enable a robot to safely
navigate in a dense crowd. The trajectories of the robot and the
humans are jointly predicted with a hand-crafted potential term to
model interactions.
Because the potential term is hand-crafted, it is possible that the
robot trajectories generated may not resemble socially compliant human
behavior.  In this paper, we learn the interaction model from
real-world pedestrian trajectories in order to predict human-like
trajectories.

In safety-critical applications like robotics, where robots are
expected to navigate safely among humans, we need to account for
uncertainty in our predictions. Those approaches that learn
interaction models but that do not deal with uncertainty as
in~\cite{alahi16} can lead to over-confident predictions which could
result in awkward and disruptive behavior. Our approach considers the
uncertainty regarding intentions (or goals) of the pedestrians and
results in accurate predictions.

The summary of our contributions in this paper are as follows: We
develop a new algorithm for enabling robots to move naturally through
dense crowds.
Following the key insight that agents in crowd engage in \textit{joint
  collision avoidance}, we develop an approach that models the
distribution over future trajectories of all interacting agents
through a joint density model that captures the idea of
\textit{cooperative} planning, i.e., agents cooperating with each
other to achieve their goals and avoid collisions.  To capture
collision avoidance behavior, we learn a local interaction model that
encodes how agents move based on how populated their vicinity is from
real human trajectory data.  During prediction, our model infers the
goal of an agent from its past trajectory and uses the learned model
to predict its future trajectory.  Finally, we demonstrate that our
model is capable of predicting robot trajectories that are natural and
human-like by reporting the experimental results on the ETH pedestrian
video dataset~\cite{pellegrini09}.

In the remainder of this paper, we will give an overview of relevant
existing work in section \ref{sec:related-work}. The notation and the
problem is defined in section \ref{sec:problem-definition}. Section
\ref{sec:approach} will describe our approach in detail and its
evaluation on a real world dataset is presented in section
\ref{sec:evaluation}. The results of the evaluation are discussed in
section \ref{sec:discussion} and the conclusions are presented in
section \ref{sec:conclusion}, along with directions for future work.

\section{RELATED WORK}
\label{sec:related-work}

\subsection{Navigation in Uncertain Dynamic Environments}
\label{sec:navig-dynam-envir}
Common approaches to robot navigation in uncertain, populated environments typically compute a path to the goal without taking into account the collaborative collision avoidance behavior exhibited by humans. Most of the approaches instead rely on local reactive collision avoidance methods \cite{siegwart03, thrun99}. Although these methods effectively avoid collisions with humans, the planned trajectories are usually suboptimal, compared to the trajectory a human would take for the same start and goal, and not socially compliant due to evasive maneuvers.

Naively modeling unpredictability in the trajectories of humans, using linear models like Kalman filters, leads to increasing uncertainty that makes safe and efficient navigation difficult \cite{trautman10}. Several works have focused on controlling the uncertainty in these predictive models by developing accurate human motion models \cite{thompson09, bennewitz05}, but these approaches do not account for interactions between humans and thus cannot model joint collision avoidance. 

\subsection{Modeling human interactions}
\label{sec:inter-betw-humans}

The social forces model proposed by \cite{helbing95} models motion of pedestrians in terms of forces that drive humans to reach a goal and to avoid obstacles. Subsequently, approaches have been proposed that use learning methods to fit the social forces model to observed crowd behavior \cite{helbing09, johansson07}. The social forces model has been shown to perform well in predicting trajectories in simulated crowds, but fails when it comes to predicting movements of pedestrians in real dense crowds as it uses a hand-crafted potential term based on distances, and doesn't learn human-human interactions from real data. Using a hand-crafted potential term results in a model that can capture simple interactions, like repulsion and attractions, but may fail to account for more complex crowd behavior. \cite{adrien06} models pedestrian motion behavior using dynamic potential fields that guide people to avoid collisions and move towards the goal, but this model does not explicitly account for interactions like cooperation between agents in the crowd.

Hall \cite{hall63} introduces a theory on human proximity relationships which have been used in potential field methods to model human-human interactions \cite{svenstrup10, pradhan11}.
These models capture interactions to avoid collisions, but do not model human-robot cooperation. However,  models of cooperation are necessary for safe and efficient navigation in dense crowds \cite{trautman10}, because in cases where the crowd density is high, the robot believes there is no 
feasible
path in the environment unless it accounts for cooperation from the crowd. Reciprocal Velocity Obstacles (RVO), \cite{vandenberg08}, and Velocity Obstacles (VO), \cite{fiorini98}, account for interactions between agents by computing joint collision-free velocities assuming constant velocities and shared collision avoidance behaviors. However, these approaches cannot handle stochastic behavior in pedestrian motions and do not train the model from real observed data. 

Trautman et. al. \cite{trautman10} proposed \textit{Interacting Gaussian processes} (IGP) to explicitly model the human-robot cooperation. Similar to the work presented in this paper, IGP models the trajectories of all interacting agents jointly which results in a probabilistic model that can capture joint collision avoidance behavior. However, the IGP model assumes that the final destinations of all pedestrians are known, which is not the case in a realistic prediction task. 
Another drawback of IGP is the use of hand-crafted interaction potential term to model cooperative behavior which may result in robot trajectories that are not socially compliant. In this paper, we learn the interaction model from observations of real pedestrian trajectory data in the hope that we achieve more human-like and socially compliant trajectories.

The works of \cite{kuderer12, Kretzschmar16} are also closely related to our work. These approaches explicitly model human-robot cooperation and jointly predict the trajectories of all agents, using feature-based representations. 
Unlike our proposed approach, they use \textit{maximum entropy inverse reinforcement learning} (IRL) to learn an interaction model from human trajectory database using carefully designed features 
such as clearance, velocity, or group membership.
However, their approach has been tested in scripted environments with no more than four humans. In our work, we deal with crowded scenarios with an average of six humans in a single scene. Very recently, \cite{pfeiffer16} have extended the maximum entropy approach to unseen and unstructured environments by using a receding horizon motion planning approach.

\subsection{Trajectory prediction}
\label{sec:traj-pred}

A large body of works exist in the domain of computer vision and video surveillance that deal with predicting motion of people in videos, that are relevant to our work. \cite{kim11, joseph11} learn motion patterns using Gaussian processes and cluster human trajectories into these motion patterns. But these approaches ignore human-human interactions. IRL has also been used in the domain of activity forecasting to predict human paths in static scenes, \cite{kitani12} and more recently, \cite{alahi16} used Long Short-Term Memory networks (LSTM) to jointly reason across multiple agents to predict their trajectories in a scene. However, most of these approaches have been used in the context of prediction and have not been extended to navigation. 

\section{PROBLEM DEFINITION}
\label{sec:problem-definition}

\subsection{Notation}
\label{sec:notation}

We follow the notation of \cite{trautman10}.
Let the index $i \in \{1,2,\cdots,N\}$ specify the agent, where $N$ is the number of individuals in the crowd and $i=R$ indicates the robot. 
The trajectory of agent $i \in \{R,1,2,\cdots,N\}$ is given by $\fbold\supi$ $= (f_1^{(i)}, f_2^{(i)}, \cdots, f_T^{(i)})$, where $T$ is the length of the trajectory and $f_t^{(i)} = (x_t^{(i)}, y_t^{(i)}) \in \mathbb{R}^2$ is the location of agent $i$ at time-step $t$. The observed locations of pedestrian $i$ until time-step $t$ is denoted as $\zbold\supi\subt = (z_1^{(i)}, z_2^{(i)}, \cdots, z_t^{(i)})$.  We denote the set of all pedestrian trajectories by $\fbold=\{\fbold\supi\}_{i=1:N}$, the robot trajectory by $\fbold\supR$, and the set of all pedestrian observations until time-step $t$ by $\zbold\subt = \{\zbold\supi\subt\}_{i=1:N}$. We assume a fixed number of goals $g$ in the environment are known and denote the set of goals by $\Gbold$. %

\subsection{Planning using the joint density}
\label{sec:planning}
Following the assumption that people engage in a joint collision
avoidance when moving through a dense crowd
 as in \cite{helbing95,trautman10},
the robot does not only have to respond to the observed trajectories
of the pedestrians, but also has to account for the adaptive behavior
of the humans.
To capture this cooperative behavior, it has been suggested by
~\cite{trautman10} to use the joint density of both the robot and the
crowd, denoted by $P(\fbold\supR, \fbold|\zbold\subt)$.
Planning the path for the robot using this density corresponds to
finding the maximum-a-priori (MAP) assignment for the following
posterior:
\begin{equation}
  \label{eq:8}
  (\fbold\supR, \fbold)_* = \argmax_{\fbold\supR, \fbold} P(\fbold\supR, \fbold|\zbold\subt).
\end{equation}

\subsection{Problem}
\label{sec:problem}
In this work, we assume that at each time-step $t$, we receive the observation $\zbold_t$ of the locations of all agents in the crowd. Given the current and past observations $\zbold\subt$, we tackle the problem of estimating the joint posterior distribution of the future trajectories of all agents. Formally, we seek to model the density given by,
$$ P(\fbold\supR, \fbold | \zbold\subt).$$
Planning the robot's trajectory then corresponds to taking the MAP assignment for $\fbold\supR$ and executing it until the next observation is received. At time-step $t+1$, we receive a new observation $\zbold_{t+1}$ and update the above joint posterior density to $P(\fbold\supR, \fbold|\zbold_{1:t+1})$. This process is repeated until the robot reaches its destination. 
In contrast to \cite{trautman10}, who tackle a similar problem, we aim to predict more natural and human-like robot trajectories by learning the model from pedestrian trajectory data. 

\section{APPROACH}
\label{sec:approach}

\begin{figure}[t!]
  \centering
  \includegraphics[width=\linewidth]{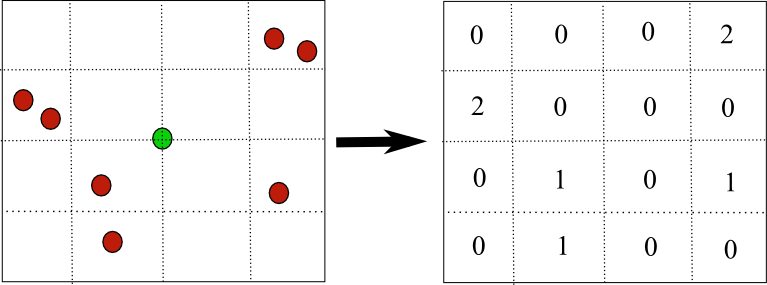}
  \caption{Occupancy grid construction. (Left) A configuration of
    other agents (red) around the current agent (green). (right) 4x4
    occupancy grid is constructed using the number of agents in each
    grid cell}
  \label{fig:ogrid}
  \vspace{-15pt}
\end{figure}

We exploit the observation that humans navigating in dense crowds
adapt their trajectories based on the presence of agents in their
vicinity \cite{alahi16}.
We first explain the construction of \textit{occupancy grids}, which
account for the presence of other agents within an agent's local
neighborhood (Section \ref{sec:occupancy}).  We formulate the problem
of learning the social interaction model as a Gaussian process
regression problem, where we predict the agent's velocities at each
time-step as a function of their occupancy grids and intended
goal. Given the preprocessed training trajectories (Section
\ref{sec:preprocessing}), we train the GP model by maximizing its
marginal likelihood to learn the hyperparameters of the kernel
(Section \ref{sec:train}). At prediction time, we use the learned
model to infer the goal of each agent and jointly predict future
trajectories of all interacting agents in the crowd (Section
\ref{sec:prediction}).

\subsection{Constructing occupancy grids}
\label{sec:occupancy}
To capture the local interactions of an agent with its neighbors, we
construct an occupancy grid for each agent $i$ at each time-step $t$
that is similar to the social pooling grid used in \cite{alahi16}. The
occupancy grid is constructed by discretizing the neighborhood of the
agent's current location into a spatial grid of size $M \times M$. We
then count the number of surrounding agents within each grid
cell. Formally, we define occupancy grid as a vector of length $M^2$
given by,
\begin{equation}
  \label{eq:2}
  \Obold_t\supi (a + M(b-1)) = \sum_{j \in \mathcal{N}\supi}\mathbb{I}_{a,b}[x_t\supj - x_t\supi, y_t\supj - y_t\supi].
\end{equation}
$\mathcal{N}\supi$ denotes the set of all agents $j \neq i$, who are
within the neighborhood of agent $i$. The indicator function
$\mathbb{I}_{a,b}[x,y]$ defines if $(x,y)$ is located in the $(a,b)$
cell of the occupancy grid.
In the remainder of the paper, $\Obold\supi$ and $\Obold$ will denote
the set of all occupancy grids at every time-step of an agent $i$ and
that of all agents, respectively.
\subsection{Preparing training data for learning}
\label{sec:preprocessing}

Before training our model, we preprocess the training data $\fbold$,
which correspond to the trajectories of all agents. First, we
construct occupancy grids $\Obold\supi$ and $\Obold$ along $\fbold$, as described in
Section \ref{sec:occupancy}. Second, we process all the trajectories
to obtain the velocities of agents at each time-step $(\xvel,
\yvel)$. Third, since we have access to the entire trajectory at
training, we can compute the goals of all the pedestrians.  Note that
this information about the true goals is used only during training and
is not assumed in the prediction phase.  After this preprocessing, we
have $(\Obold, \xbvel, \ybvel)$ for all agents at every time-step and
their corresponding goals $\{g\supi\}$.%

\subsection{Training the local interaction model}
\label{sec:train}

To learn across different pedestrians traversing in different regions
of the environment, we model the velocity of an agent at a specific
time-step as a function of their intended goal and their occupancy
grid at that time-step. Formally, we seek to estimate the distribution
$P(\xbvel| \Obold, g)$ and $P(\ybvel| \Obold, g)$, for each goal $g$
in $\Gbold$, from the training data obtained in Section
\ref{sec:preprocessing}.

We start by modeling the interactions as a Gaussian Process (GP)
regression problem, where the noisy data to be interpolated is
$(\Obold, \xbvel)$ and $(\Obold, \ybvel)$. Note that we use a separate
GP for each goal $g$ to learn the mapping from occupancy grids to
velocities. We use a \textit{squared exponential automatic relevance
  determination} (SE-ARD) kernel (with additive noise) for these GPs
\cite{rasmussen06}.  The SE-ARD kernel learns a different lengthscale
for each input dimension, which in our problem are the dimensions of
the occupancy grid vector, i.e., the grid cells. Since, the kernel can
learn the relevance of each input dimension by learning the
lengthscale parameter \cite{rasmussen06} (dimensions with large
lengthscales are less relevant than those with small lengthscales), the
kernel will capture the relevance of each grid cell for a specific
goal and effectively ignore irrelevant cells.

The SE-ARD kernel with additive noise is given by,
\begin{align}
  \label{eqn:kernel}
  \begin{split}
    K_{S}(\Obold, \Obold') &= \sigma_f^2 \exp\left(-\frac{1}{2}\sum_{d=1}^{M^2} \frac{(\Obold(d) - \Obold'(d))^2}{\ell_d^2}\right) \\
    &+ \sigma_n^2 \delta(\Obold, \Obold')
  \end{split}
\end{align}
where $\delta(\Obold,\Obold') = 1$ if $\Obold$ is equal to $\Obold'$
and zero otherwise, and $\Obold(d)$ is the value of the $d^{th}$
dimension in the vector $\Obold$. The hyperparameters of this kernel
are $\sigma_f$ (signal variance), $\{\ell_d\}_{d=1}^{M^{2}}$
(lengthscales) and $\sigma_n$ (noise variance).

This construction results in a total of $2G$ GPs because there is a
pair of GPs (in x- and y-direction) for each of the $G$ goals; thus,
we have $2G$ sets of hyperparameters to be learned. We denote the set
of hyperparameters corresponding to the GP associated with goal $g$ by
$\Theta_x^g$ and $\Theta_y^g$. To learn the hyperparameters
$\Theta_x^g$, we isolate the tuples
$B_g = \{(\Obold\supi, \xbvel\supi)\}_i$ from training data,
corresponding to the set of pedestrians $i$ whose goal is $g$, and
maximize the log marginal likelihood of the GP \cite{rasmussen06}
given by,

\begin{align}
  \label{eqn:logmarginal}
  \begin{split}
    \log P\left(\xbvel|\Obold\right) &= -\frac{1}{2}\xbvel^TK_S(\Obold, \Obold)^{-1}\xbvel \\
    &- \frac{1}{2}\log|K_S(\Obold, \Obold)| - \frac{n_g}{2}\log 2\pi
  \end{split}
\end{align}
where $\xbvel$ and $\Obold$ are vectors constructed by concatenating
elements of $B_g$, and $n_g$ is the number of elements in $\xbvel$.

We can learn $\Theta_y^g$ and all the other sets of hyperparameters
for all goals $g \in \Gbold$ in a similar fashion.

\subsection{Prediction}
\label{sec:prediction}

During prediction we are given an unseen crowd with $N_p$ pedestrians,
a robot, and their observations $\zbold_{1:t}$ until time $t$. Our
task is to predict their trajectories $\fbold$ and $\fbold\supR$ for
$H$ time-steps into the future, using the learned model. We cannot
directly use the GP predictive distribution because we do not know the
goals of the pedestrians during prediction. Note that we know the goal
of the robot $g\supR$ since it is user-defined.

\subsubsection{Infer goal of a pedestrian}
\label{sec:infer}

Given observations $\zbold\supi\subt$ of pedestrian $i$ until time $t$
and the set of goals $\Gbold$, we seek to infer the goal
$g\supi \in \Gbold$ of the pedestrian. We assume a uniform prior
$P(g\supi)$ over all goals, in the absence of any observations for
agent $i$ (A more informative prior over the goals can be found by
analyzing the environment). Hence, we have:
\begin{equation}
  \label{eq:1}
  P(g\supi|\zbold\supi\subt) = \frac{P(\zbold\supi\subt|g\supi) P(g\supi)}{P(\zbold\supi\subt)} \propto P(\zbold\supi\subt|g\supi).
\end{equation}

That is, we evaluate the likelihood that the observation sequence
$\zbold\supi\subt$ is true conditioned on the fact that $g\supi$ is
the goal of agent $i$. Similar approaches have been explored in
\cite{kitani12} and \cite{ziebart09}, for inferring destination of an
agent given its previous path.

To compute the likelihood, we first compute, from the observations
$\zbold\subt$, the velocities $\{(\xbvel, \ybvel)\}_{1:t-1}$ and the
occupancy grids $\Obold_{1:t-1}$ of all pedestrians at each time-step
until $t-1$.  For each possible goal $g \in \Gbold$, we take the
corresponding set of trained hyperparameters $\Theta_x^g$ and
$\Theta_y^g$, and evaluate the log marginal likelihood of the GP for
each agent $i$ (using equation \ref{eqn:logmarginal}).
Hence, for each agent $i$ and each goal $g\supi \in \Gbold$, we obtain
the likelihood that its observed data $\zbold\supi\subt$ is generated
from the GP conditioned on the goal $g\supi$. Normalizing the
likelihoods across all goals, we get the likelihood
$P(\zbold\subt\supi|g\supi)$ for every goal $g\supi \in \Gbold$.

\subsubsection{Predicting future trajectories}
\label{sec:pred-future-traj}

Now that we have a distribution over the goals $g\supi$ for all agents
$i$ in the crowd, we can use the trained model to predict future
locations. The joint posterior density can be decomposed as
\begin{equation}
  \label{eq:11}
  P(\fbold\supR, \fbold | \zbold\subt) = \sum_{\gbold} P(\fbold\supR, \fbold | \gbold, \zbold\subt) P(\gbold | \zbold\subt)
\end{equation}
where $\gbold = \{g\supi\}_{i=R,1:N}$ are the goals of all agents
including the robot.
We can assume that the goals of the pedestrians are independent of
each other (and that we know the goal of the robot with certainty)
given their respective observations. Then, we can write the
distribution of a goal given a history of observations as:
\begin{equation}
  \label{eq:12}
  P(\gbold|\zbold\subt) = \prod_{i=1}^{N} P(g\supi|\zbold\subt\supi)
\end{equation}
where $P(g\supi|\zbold\subt\supi)$ is given by equation \ref{eq:1}.
We approximate the joint distribution
$P(\fbold\supR, \fbold|\gbold, \zbold\subt)$ by using the velocities
and occupancy grids obtained from observations $\zbold\subt$ (as done
in Section \ref{sec:infer}),
\begin{equation}
  \label{eq:13}
  P(\fbold\supR, \fbold|\gbold, \zbold\subt) \approx P(\fbold\supR, \fbold|\{\xbvel, \ybvel\}_{1:t-1}, \Obold\subt, \gbold)
\end{equation}
The predictions for different agents are coupled through the occupancy
grid which contains the configuration of other agents around each
agent locally.
This enables our model to capture local interactions, like joint
collision avoidance and cooperation.

Since the task is to predict the future locations of all agents for
the next $H$ time-steps, $\fbold\supi$ suffices to represent the next
$H$ locations of agent $i$ after time $t$, in addition to previous
locations.

\subsubsection{Multi-step prediction}
\label{sec:sampling}
Future locations can be predicted using the learned model from Section
\ref{sec:train}. For each agent $i$, we fit a separate pair of GPs
(with the learned hyperparameters $\Theta_x^g$, $\Theta_y^g$ for goal
$g$
) to the observed tuples $(\Obold\supi\subt, \{\xbvel\}\subti\supi)$
and $(\Obold\supi\subt, \{\ybvel\}\subti\supi)$. Using their
corresponding GPs, each agent can predict their velocities and compute
the location for the next time-step by adding it to the current
location.

This can be done exactly for time $t+1$, i.e., we can predict
$(\xbvel)_{t+1}\supi$ and $(\ybvel)_{t+1}\supi$ for each agent $i$,
since we know the value of the occupancy grid at time $t$,
$\Obold\supi_t$. But for future time-steps, we need to estimate the
occupancy grid at the previous time step using previous
predictions. Instead of computing the distribution over future
locations in an exact form (which can be extremely difficult), we use
Monte Carlo sampling to approximate the distribution as shown in
Algorithm \ref{alg:sampling}.

At time $t+1$, we compute the GP predictive distribution
\cite{rasmussen06} for each $(\xbvel)_{t+1}\supi$ and
$(\ybvel)_{t+1}\supi$ (line \ref{alg:line:start}).
We then proceed to sample $S$ points from each of these distributions
(line \ref{alg:line:sample}), and estimate $S$ samples for the
location $\fbold_{t+1}\supi$ for all agents $i$ (line
\ref{alg:line:estimate}).  These sets of samples approximate the
distribution
$P(\fbold\supR_{t+1}, \fbold_{t+1}|\{\xbvel, \ybvel\}_{1:t-1},
\Obold\subt, \gbold)$.

Since, for each sample, we have locations of all agents at time $t+1$,
we can compute occupancy grids $\Obold\supi_{t+1}$ for each agent
(line \ref{alg:line:estimateocc}). Thus, we get $S$ samples for
$\Obold_{t+1}$.  Now, to estimate location at time $t+2$, we compute
the mean of the $S$ samples to get the set of occupancy grids,
$\Obold_{t+1}$ (line \ref{alg:line:mean}).  Using the mean,
we predict the velocities at time $t+2$ (line \ref{alg:line:repeat}),
and repeat the above process until $H$ time-steps into the future. At
every time-step $t'$ ($t' \geq t+1, t' \leq t+H$), we get a set of $S$
samples corresponding to the locations $\fbold_{t'}$ that approximate
the distribution
\begin{align*}
  \{\fbold\supR_{t'}, \fbold_{t'}\}_{j=1:S} \approx P(\fbold\supR_{t'}, \fbold_{t'}|\{\xbvel, \ybvel\}_{1:t-1}, \Obold\subt, \gbold)
\end{align*}
As we let the value of $S$ grow, we get a better approximation.

\begin{algorithm}[h]
  \begin{algorithmic}[1]
    \For{each agent $i$} \State Compute distributions of
    $(\xbvel)_{t+1}\supi$, $(\ybvel)_{t+1}\supi$ given
    $\Obold\supi_t$\label{alg:line:start}
    \EndFor
    \For{$t'=t+2 \to t+H$} \For{each agent $i$} \State Sample $S$
    points from distributions of $(\xbvel)_{t'-1}\supi$,
    $(\ybvel)_{t'-1}\supi$\label{alg:line:sample} \State Compute $S$
    estimates for $\fbold\supi_{t'}$ from sampled
    velocities\label{alg:line:estimate}
    \EndFor
    \State Compute $S$ samples for $\Obold_{t'}$ from estimates of
    $\fbold_{t'}$\label{alg:line:estimateocc} \State Set $\Obold_{t'}$
    to be the mean of the $S$ samples from above \label{alg:line:mean}
    \For{each agent $i$} \State Compute distributions of
    $(\xbvel)_{t'}\supi$, $(\ybvel)_{t'}\supi$ given
    $\Obold\supi_{t'}$\label{alg:line:repeat}
    \EndFor
    \EndFor \\
    \Return
    $\{\{\fbold\supR_{t'},
    \fbold_{t'}\}_{j=1:S}\}_{t'=t+1:t+H}$\label{alg:line:return}
  \end{algorithmic}
  \caption{Multi-step prediction through Sampling}
  \label{alg:sampling}
\end{algorithm}

\section{EVALUATION}
\label{sec:evaluation}

\subsection{Setup}
\label{sec:setup}

\begin{figure}[t!]
  \centering
  \includegraphics[scale=0.4]{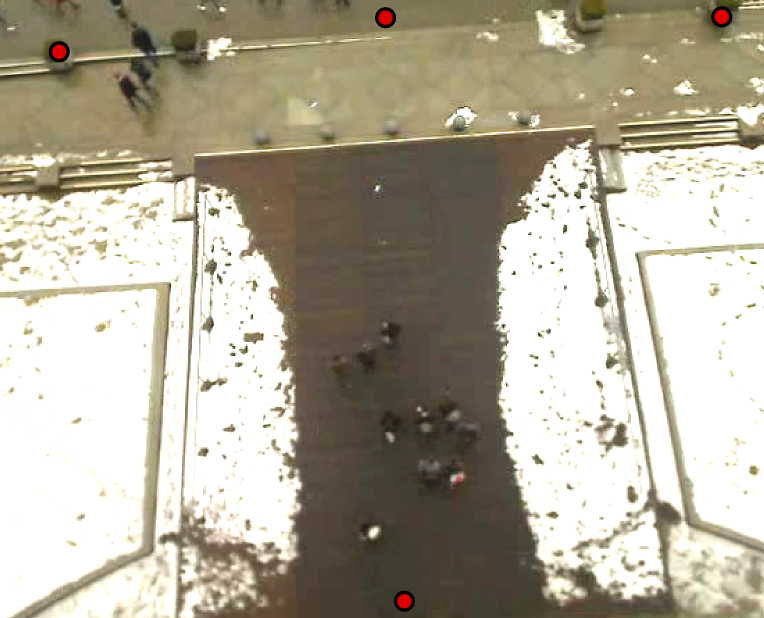}
  \caption{Example snapshot of the dataset with goals indicated by red
    dots}
  \label{fig:snapshot}
  \vspace{-15pt}
\end{figure}

We evaluate our model on a publicly available human-trajectory dataset
released by ETH, \cite{pellegrini09}.
The dataset contains a video recorded from above a busy doorway of a
university building with the pedestrian trajectories tracked and
annotated.  This video contains scenes with real world crowded
settings, hundreds of trajectories and high crowd density.
An example snapshot from the video (with goals marked) is shown in
figure \ref{fig:snapshot}. Each time-step in the video is six frames
long and amounts to 0.4 seconds. The average trajectory length of a
pedestrian is 25 time-steps. The total number of pedestrians in the
video is 360 and there are four goals in the environment. The
resolution of the video is $640 \times 480$. Each pixel in the video
frame corresponds to 0.042 metres (slightly varies across the frame,
as the camera is angled and not exactly top-down).

We evaluate our model by choosing a pedestrian in the crowd randomly
as our robot and use his start and goal state to plan a path through
the crowd. The resulting path is compared to the true path that the
pedestrian has taken in the video. Comparing the true path and the
predicted path gives us a evaluation of how closely our prediction
resembles human-like behavior. A similar evaluation was done in
\cite{trautman10}.

We compare our approach against IGP \cite{trautman10}, as it deals
with the same problem of jointly modeling trajectories of robot and
pedestrians, and has shown good results in real robot navigation
scenario among dense crowds \cite{trautman13, trautman15}.
We will use both our approach and IGP to predict the path until $H$
time-steps into the future and compare it with the true trajectory.

Note that the original IGP needs to know the true final destination of
each pedestrian at prediction time, which would give it an unfair
advantage over our algorithm. Hence, we use a variant of IGP which
doesn't need the true final destinations and use that in our
comparison. At prediction time, we compute the average heading of the
pedestrian for the last 5 time-steps and estimate the goal location in
the computed heading direction. The estimated goal is used in the
original IGP algorithm in place of the true final destination of the
pedestrian.

To compare the path predicted by the two algorithms and the true path
of the pedestrian, we consider two metrics:
\begin{enumerate}
\item \textit{Average displacement error}: Introduced in
  \cite{pellegrini09}, this metric computes the mean squared error
  over all estimated points at each time-step in the predicted
  trajectory and the true trajectory.
\item \textit{Final displacement error}: Introduced in \cite{alahi16},
  this metric computes the mean distance between the final predicted
  location after $H$ time-steps and the true location after $H$
  time-steps, where $H$ is our prediction horizon.
\end{enumerate}

\subsection{Model Parameters}
\label{sec:model-parameters}

We construct occupancy grids around each agent of size $4 \times 4$ (i.e. $M=4$) covering a space of $80 \times 80$ pixels in the video. In each video, we train the model on the trajectories of the first 50 pedestrians. At prediction time, we chose a scenario with 11 pedestrians (from the remaining part of the video, not used in training), one of whom is used as a robot in our model. We predict the future locations for a range of prediction horizons $H = 1,2,5,10,20$ time-steps. If the path of the pedestrian (or robot) ends in less than $H$ time-steps, we will predict only until the end of his path. For multi-step prediction, we use $S = 100$ samples to approximate the distribution over future locations. We implement the Gaussian process regression model using the GPML toolbox, \cite{rasmussen06}.

\subsection{Results}
\label{sec:results}

\begin{figure}[t!]
  \centering
  \includegraphics[scale=0.35]{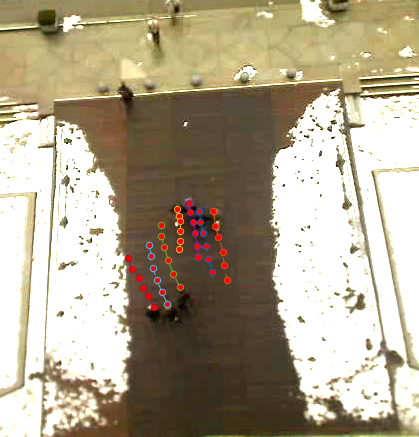}
  \caption{Example prediction by our model. For each pedestrian, we predict his future locations (which are plotted) for the next 5 time-steps. The bottom set of pedestrians are progressing towards a goal at the top centre of the image, but they go around the other set of pedestrians making way for them cooperatively}
  \label{fig:bestcase}
\end{figure}

To test the prediction accuracy of both approaches, we have chosen scenarios with 11 pedestrians in the crowd where crowd density is high and some pedestrians head into and through the crowd. To get an unbiased estimate, we chose 5 such scenarios in the video. In each scenario, we assume each pedestrian to be the robot, one at a time, and compute their average displacement error and final displacement error, averaged over all time-steps. This results in 11 sets of error values and we compute the average over all sets to give the mean errors over all pedestrians. We repeat the experiment for different $H$ values to get both short-range and long-range prediction accuracies of both approaches. The results are averaged over all 5 scenarios and are presented in Table \ref{table:results}. Note that the errors are listed in pixels.

\begin{table}[t!]
  \caption{Prediction errors (in pixels) on the dataset for IGP and our approach}
  \centering
  \begin{tabular}{|c|c|c|c|}
    \hline
    Metric & Prediction horizon ($H$) & IGP & Our Approach \\
    \hline
    \multirow{5}{*}{Avg. Disp. Error} & 1 & 3.42 & 4.42 \\
    & 2 & 5.66 & 6.14 \\
    & 5 & 15.75 & 12.09 \\
    & 10 & 21.59 & 21.52 \\
    & 20 & 41.51 & 34.63 \\
    \hline
    \multirow{5}{*}{Final Disp. Error} & 1 & 3.42 & 4.42 \\
    & 2 & 7.12 & 7.78 \\
    & 5 & 23.18 & 19.77 \\
    & 10 & 38.75 & 36.25 \\
    & 20 & 67.41 & 54.2 \\
    \hline
  \end{tabular}
  \label{table:results}
\end{table}

\begin{figure}[t!]
  \centering
  \includegraphics[width=\linewidth]{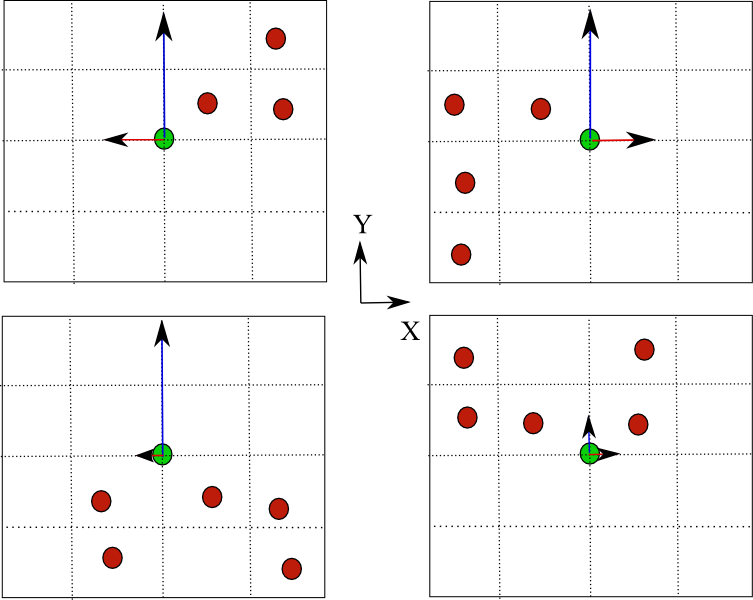}
  \caption{Velocities predicted by our trained model for example occupancy grids. In each case, the goal of the pedestrian is right above in the Y-direction. Predicted mean y-velocity is shown in blue and predicted mean x-velocity is shown in red.}
  \label{fig:learnedmodel}
  \vspace{-15pt}
\end{figure}

In Figure \ref{fig:bestcase}, we show an example scene where our approach predicts cooperative behavior. The set of pedestrians going up give way to the set of pedestrians going down. To visualize what our local interaction model (from section \ref{sec:train}) learned, we give it some example occupancy grids and goals, and observe the predicted velocities. Figure \ref{fig:learnedmodel} shows that the model learns collision avoidance as it predicts velocities away from grid cells which are occupied and towards unoccupied grid cells. When an agent's vicinity is heavily populated in the direction of its goal, the magnitude of predicted velocity is very low, i.e., the agent moves slowly. If instead, its vicinity is populated in the opposite direction of its goal, the velocity of the agent doesn't get affected by the surrounding agents (as they are not obstructing its path).

To verify this observation, we have examined the values of the learned hyperparameters of the SE-ARD kernel. The lengthscales for grid cells that are not in the direction of the pedestrian's intended goal, are given high values, thus reducing their relevance in the velocity prediction. For example, in Figure \ref{fig:learnedmodel} the lengthscales for the bottom grid cells in the left bottom occupancy grid, are set to values higher than 7 whereas the lengthscales for the top grid cells in the same grid are set to values lower than 1. This shows that our model learns how neighbors affect a pedestrian's path based on their relative spatial orientation. 

\section{DISCUSSION}
\label{sec:discussion}

From the results presented in Table \ref{table:results}, we can
observe that our approach performs better than IGP at predicting human
trajectories for longer prediction horizons and worse for shorter
horizons. This is mainly because IGP models trajectories directly by
predicting future locations based on previously observed locations. This
results in very accurate predictions in situations where there are no
surrounding agents (and hence, no interactions) and for shorter
prediction horizons, as it extrapolates the trajectory to future
time-steps smoothly. Our model, on the other hand, models velocities
at each time-step and needs to estimate the future location based on
velocity predictions for the previous time-step. Thus, for shorter
prediction horizons, our model has a higher variance associated with
predicted locations. But for longer horizons, our model has higher
accuracy in prediction as it reasons about the intended goal of the
agent and captures local interactions at each time-step. IGP fails at
longer horizons as the smooth extrapolation, coupled with the
handcrafted interaction potential term, is unable to account for all
the interactions and cooperative behavior among dense crowds. More
importantly, our approach has a higher performance than IGP because it
learns the local interaction model from real human trajectory data
whereas the interaction model in IGP is hand-crafted.

Accurate predictions for longer horizons is important in social navigation
as it results in a more globally optimal behavior. In cases where the
prediction is accurate in a short horizon but poor for longer horizons,
the resulting paths are locally optimal and can potentially lead to a
non-socially compliant and reactive behavior.

Upon careful examination of predictions of our approach in crowded
scenarios, we observed that it learns the behavior of slowing down
(see bottom right of Figure \ref{fig:learnedmodel}) when its vicinity
is heavily populated, which is a common behavior among humans. Also,
as observed from the values of the learned lengthscales for the SE-ARD
kernel, our model learns how humans decide their velocity based on the
relative spatial configuration of other agents in their
neighborhood. As shown in Figure \ref{fig:learnedmodel}, our trained
local interaction model captures collision avoidance based on an
agent's occupancy grid by learning from human trajectory data without
any hand-crafted potential term.

Although we present results for predicting trajectories of every
agent in the crowd, this approach can be extended to robot navigation by
treating the robot as an agent in the crowd. Planning the path of the
robot in this model reduces to inference in the joint density as shown
in Section \ref{sec:planning}. The resulting path taken by the robot
is the most likely path predicted according to the learned
model. Recent work by \cite{pfeiffer16} has shown that as long as
pedestrians interact with the robot naturally (as one of them), such
an interaction-aware modeling approach is significantly better than a
reactive approach.

\section{CONCLUSION AND FUTURE WORK}
\label{sec:conclusion}

In this work, we present a new approach to modeling cooperative
behavior among humans in dense crowds. While most existing approaches
use hand-crafted models to capture interactions among humans in the
crowd, we take a data-driven approach and learn an interaction model
from real human trajectory data. We propose a nonparametric
statistical model that uses Gaussian processes to model velocities of
agents in the crowd as a function of how populated their vicinity
is. We show how our model can be used to predict future trajectories
of pedestrians and compute the path of a robot through a dense
crowd. The future trajectories are computed using a Monte Carlo
sampling approach to multi-step prediction. Lastly, the efficacy of
our approach is demonstrated by predicting trajectories of agents in a
real world pedestrian dataset. Our results show that the model
captures important aspects of human crowd behavior such as cooperative
navigation and collision avoidance.

Currently, the model doesn't account for static obstacles which play a
very important role in modeling navigation behavior. An interesting
future direction would be to explore ways to account for both the
dynamic pedestrians and static obstacles in the environment, while
predicting future trajectories. Another important drawback of our
approach is the assumption of known goals in the environment. This
restricts the generalizability of the approach to previously seen
environments and a separate model needs to be trained for a new
environment.

As a part of future work, we plan to validate and verify our approach
on a real robot placed in a dense human crowd. The task would be,
given a start and goal location, the robot should be able to navigate
safely and efficiently through the crowd. In addition to validating
our approach, we intend to tackle the drawbacks of the approach as
stated before. We are also looking to extend the model by coming up
with latent representations of trajectories that encode time-varying
information such as pedestrian's intention, planned future path and
velocities, that can be used instead of an occupancy grid in our
approach.

\section*{ACKNOWLEDGMENTS}
The authors would like to thank Peter Trautman for sharing the code
for IGP and insightful discussions, and the anonymous reviewers for
their helpful suggestions.  This work was conducted in part through
collaborative participation in the Robotics Consortium sponsored by
the U.S Army Research Laboratory under the Collaborative Technology
Alliance Program, Cooperative Agreement W911NF-10-2-0016. The views
and conclusions contained in this document are those of the authors
and should not be interpreted as representing the official policies,
either expressed or implied, of the Army Research Laboratory of the
U.S. Government. The U.S. Government is authorized to reproduce and
distribute reprints for Government purposes notwithstanding any
copyright notation herein.

\end{document}